%% file: main.tex
\documentclass[letterpaper, 10 pt, conference]{ieeeconf}  

\IEEEoverridecommandlockouts                              

\input{macros}

\usepackage[switch]{lineno}  %

\title{Safe-Planner: A Single-Outcome Replanner for Computing Strong Cyclic Policies in Fully Observable Non-Deterministic Domains}
\author{Vahid Mokhtari, Ajay Suresha Sathya, Nikolaos Tsiogkas, Wilm Decr\'e
\thanks{All authors are with the Department of Mechanical Engineering, 
Division Robotics, Automation and Mechatronics (RAM), and Flanders Make, 
KU Leuven, Belgium.
Email: mokhtari.vahid@kuleuven.be} 
}

\begin{document}

\maketitle
\thispagestyle{empty}
\pagestyle{empty}

\begin{abstract}
Replanners are efficient methods for solving non-deterministic 
planning problems. Despite showing good scalability, existing 
replanners often fail to solve problems involving a large number 
of \emph{misleading plans}, 
i.e., weak plans~that~do~not lead to strong solutions, however, 
due to their minimal lengths, are likely to be found at every 
replanning iteration.
The poor performance of replanners in such problems is due 
to their all-outcome determinization. 
That is, when compiling from non-deterministic to classical, 
they include all compiled classical operators in a single 
deterministic domain which leads replanners to continually 
generate misleading~plans.
We introduce an offline replanner, called Safe-Planner (SP), 
that relies on a single-outcome determinization to compile 
a non-deterministic domain to a set of classical domains, 
and ordering heuristics for ranking the obtained classical 
domains. 
The proposed single-outcome determinization and the heuristics 
allow for alternating between different classical domains.
We show experimentally that this approach can allow SP to 
avoid generating misleading plans but to generate weak plans 
that directly lead to strong solutions. 
The experiments show that SP outperforms state-of-the-art 
non-deterministic solvers by solving a broader range of problems.
We also validate the practical utility of SP in real-world 
non-deterministic robotic tasks.
\end{abstract}

\input{1-introduction}

\input{2-formulation}
\input{3-safe-planner}
\input{4-experiment}



\section*{ACKNOWLEDGMENT}
This work is supported by the Flanders Make SBO MULTIROB 
project at KU Leuven, Belgium.

\bibliographystyle{IEEEtran}
\bibliography{ref}

\end{document}

%% file: macros.tex
\usepackage{color}
\usepackage[noend]{algpseudocode}
\usepackage[linesnumbered,noend]{algorithm2e}
\usepackage{listings}
\usepackage{courier}
\usepackage{amsmath}
\usepackage{multirow}
\usepackage{graphicx}
\usepackage{array}
\usepackage{subcaption}
\usepackage[font={footnotesize}]{caption}
\usepackage{amsthm}
\usepackage{thmtools}
\usepackage{amssymb}
\declaretheorem[style=definition,qed={\small$\blacksquare$}]{definition}
\declaretheorem[]{theorem}
\declaretheorem[]{lemma}

\newcommand{\mc}[1]{\mathcal{#1}}

\definecolor{red}{rgb}{1.00,0.00,0.00}
\definecolor{blue}{rgb}{0.00,0.00,1.00}
\definecolor{green}{rgb}{0.4,1.00,0.0}
\definecolor{yellow}{rgb}{0.5,0.5,0.0}
\definecolor{gray}{rgb}{0.5,0.5,0.5}
\definecolor{ipython_frame}{RGB}{207, 207, 207}
\definecolor{halfgray}{gray}{0.3}

\lstdefinestyle{customlst}{
    mathescape,
    rulecolor=\color{ipython_frame},
    frame=tblr,
    frameround={t}{t}{t}{t},
    numbersep=2pt,
    numberstyle=\tiny\color{halfgray},
    captionpos=b,
    belowcaptionskip=-9pt,
    stepnumber=1,
    basicstyle=\ttfamily\fontsize{7.4}{7.6}\selectfont,
    keywordstyle=\bfseries}
\newcommand{\sans}[1]{{\fontfamily{cmss}\selectfont{#1}}}
\newcommand{\ts}[1]{{\text{\fontfamily{cmss}\selectfont{#1}}}}

\usepackage{xspace}
\newcommand{\ND}{${D=(L,O)}$\xspace}
\newcommand{\PP}{$\mc{P}=(D,s_0,g)$\xspace}
\newcommand{\SP}{{Safe-Planner}\xspace}
\newcommand{\MSP}{{Make-Safe-Plan}\xspace}

%% file: 1-introduction.tex
\section{Introduction}
\label{sec:introduction}

\begin{figure}[!t]
  \centering
  \begin{subfigure}[b]{4.3cm}
    \includegraphics[width=\linewidth]{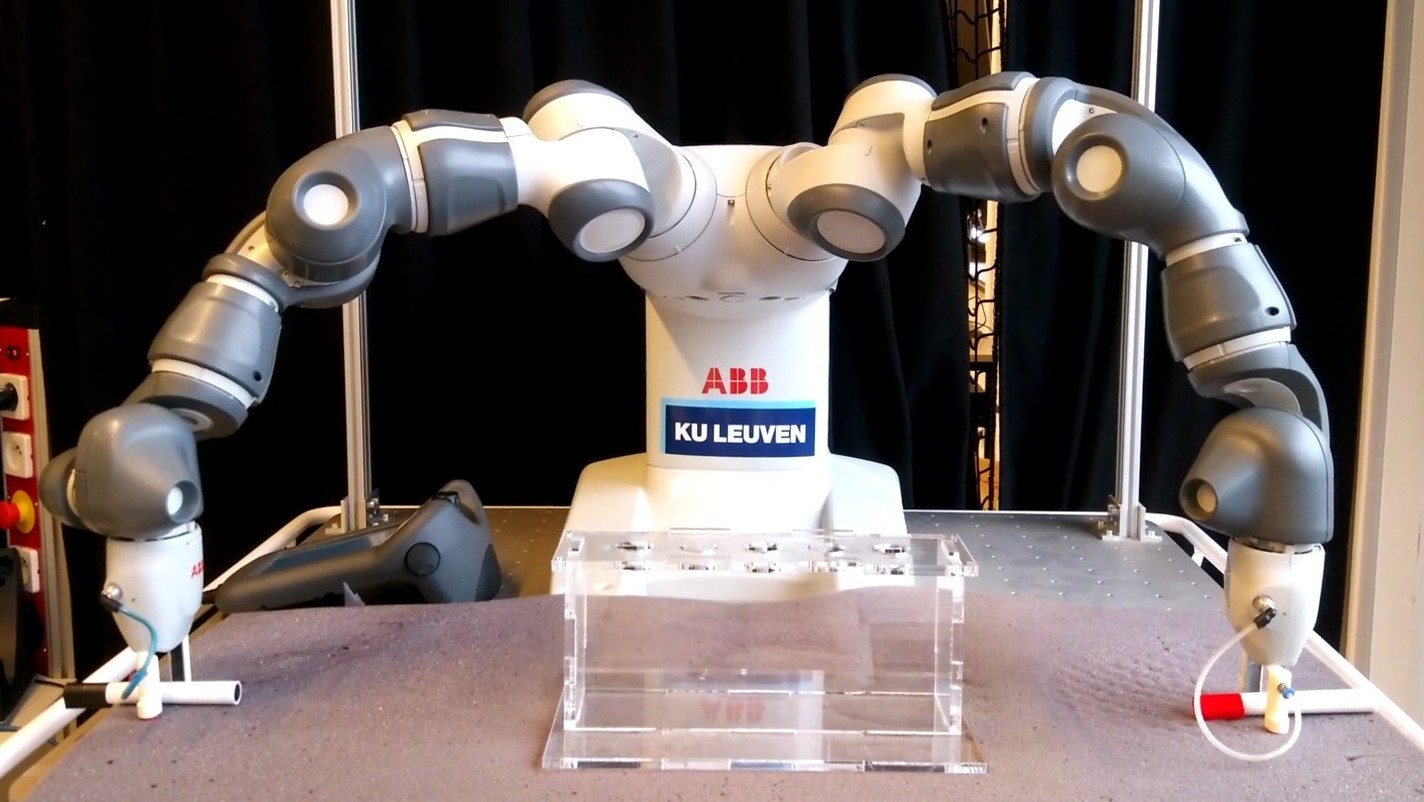}
  \end{subfigure}\hspace{-2.5pt}
  \begin{subfigure}[b]{4.3cm}
    \includegraphics[width=\linewidth]{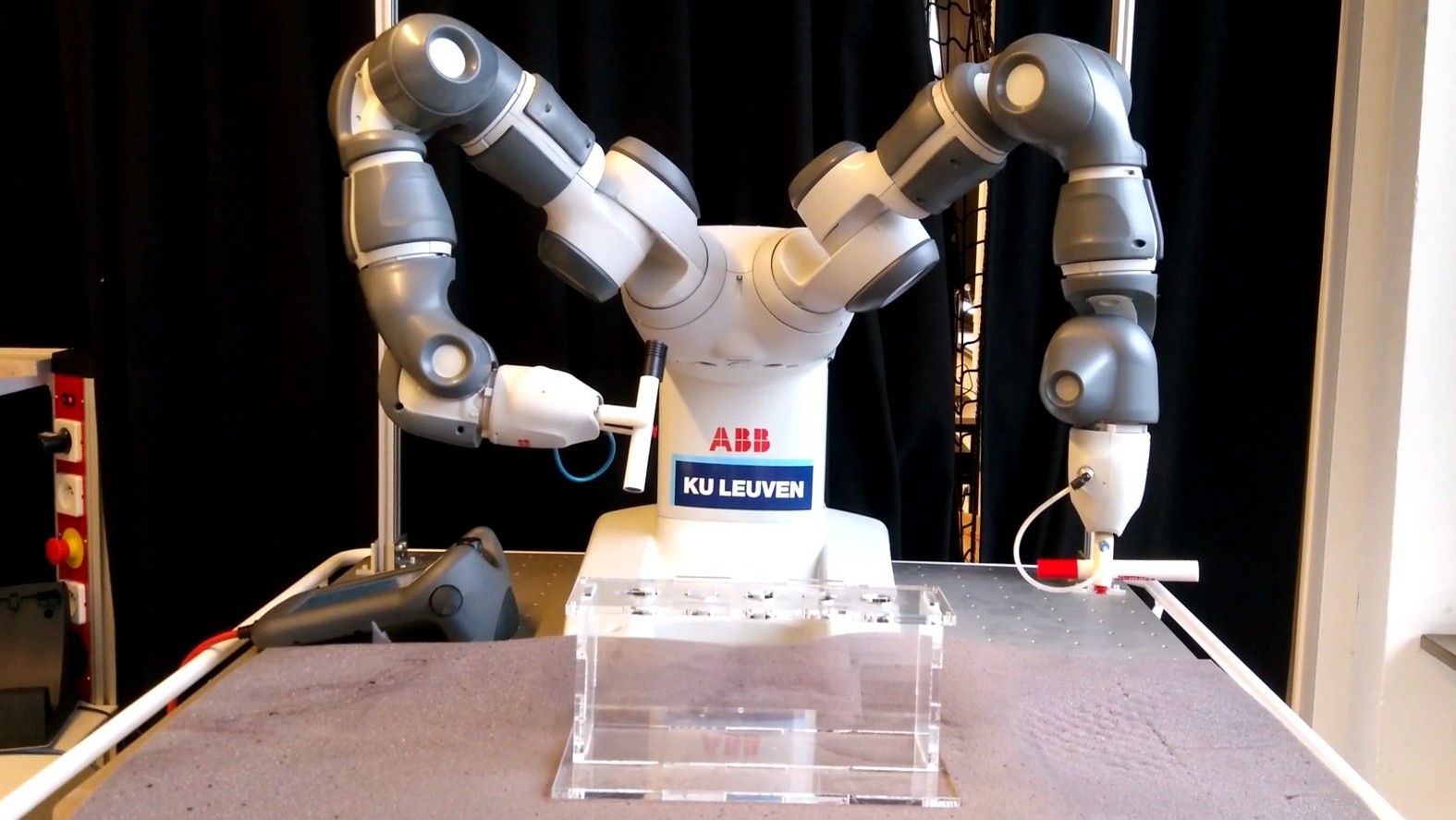}
  \end{subfigure}\\\vspace{1pt}
  \begin{subfigure}[b]{4.3cm}
    \includegraphics[width=\linewidth]{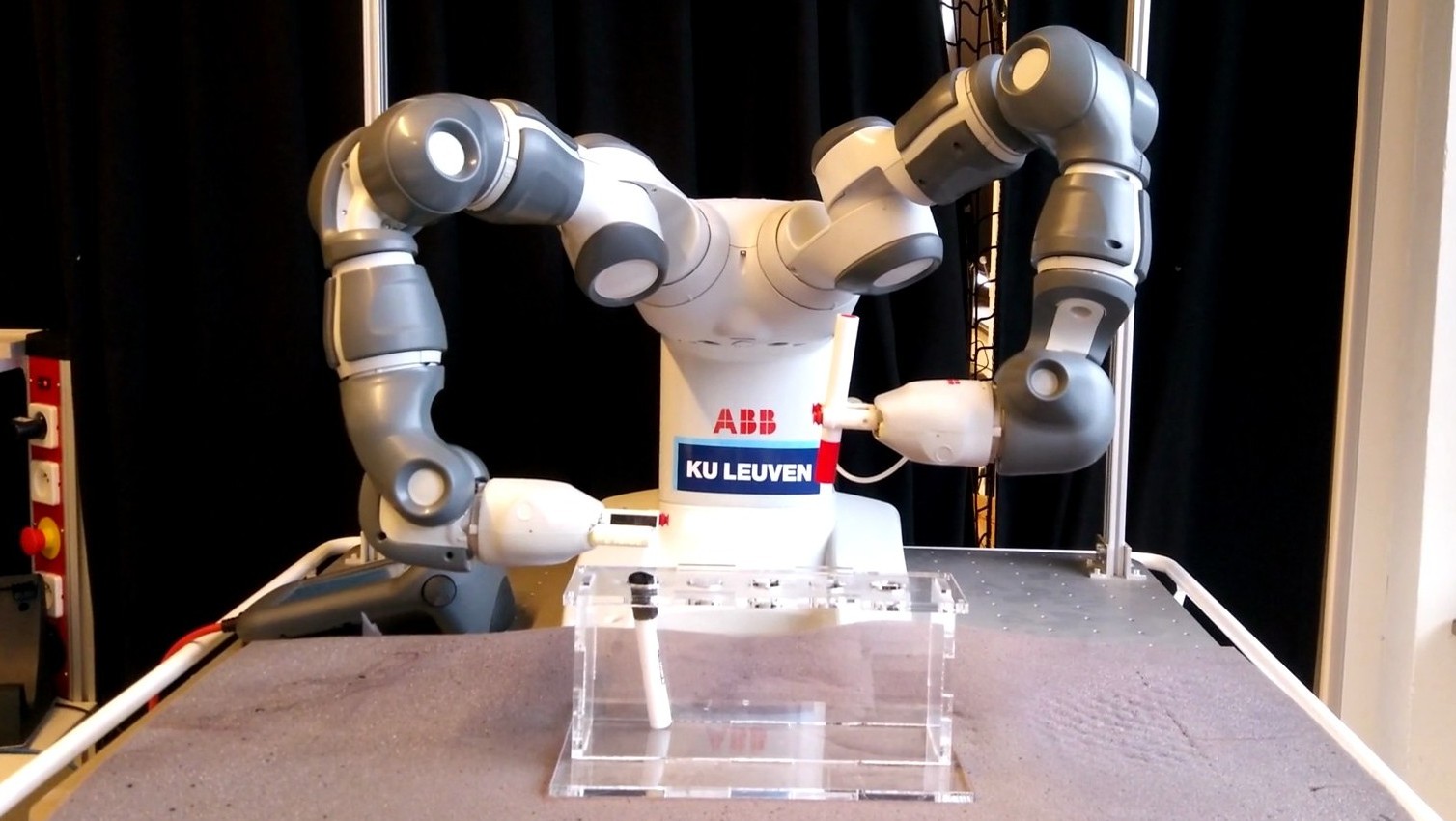}
  \end{subfigure}\hspace{-2.5pt}
  \begin{subfigure}[b]{4.3cm}
    \includegraphics[width=\linewidth]{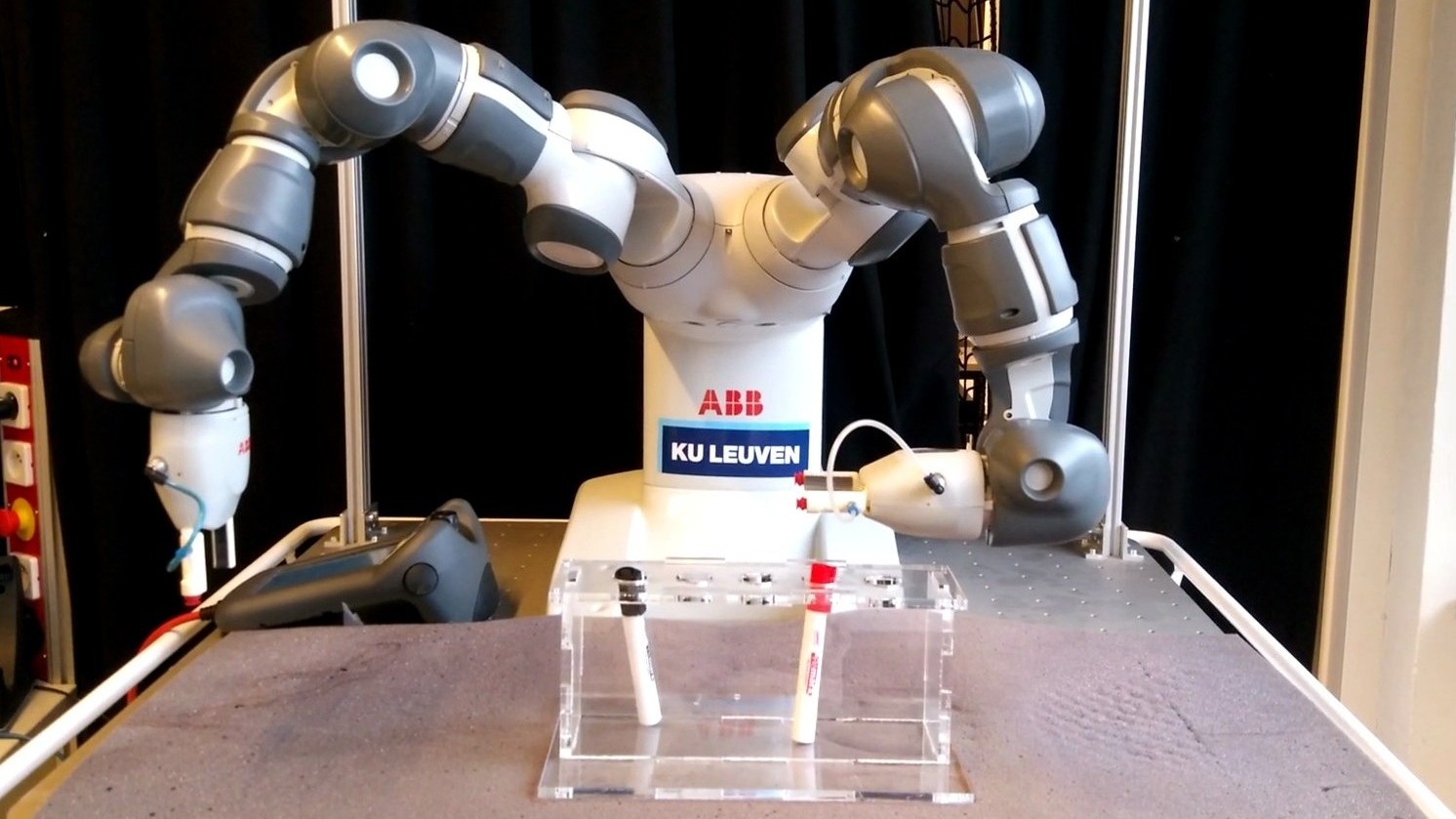}
  \end{subfigure}\\
  \vspace{-2pt}
  \caption{A dualarm YuMi manipulator puts two objects into a box 
  in an upward orientation. A strong solution, encoding contingencies,
  for the robot 
  involves picking up objects; carrying the objects to a camera; 
  checking the orientation of the objects; rotating the objects or not,
  depending on their observed orientation; and putting them into the box
  }
  \label{fig:exp:packaging}
\end{figure}

AI planning is the reasoning side of acting. It is a~core 
deliberation function for intelligent robots to increase 
their autonomy and flexibility through the construction of 
sequences of actions for achieving some prestated objectives 
\cite{ghallab2004automated}.
Major progress has been achieved in developing classical AI 
planning algorithms, however, these approaches are restricted 
to the hypothesis of deterministic actions: an applicable action 
in a state results in a single new state, thus the world evolves 
along a single fully predictable path.
A more realistic assumption must take into account that
the effect of actions can not be completely foreseen and 
the environment can change in unpredictable ways, however, 
an agent is able to observe its outcome.
As an example,~let~us consider a dualarm manipulator in 
a non-deterministic task involving picking objects from 
a table and putting them in a box in a certain orientation 
(see Figure~\ref{fig:exp:packaging}). 
The orientation of objects on the table are not known to the 
robot. In order to rotate objects in a certain orientation, 
the robot needs to observe the objects at runtime. So, a 
strong solution for the robot must be contingent on observing 
the orientation~of~objects.

We address Fully Observable Non-Deterministic (FOND) 
planning problems as an attempt at modeling uncertainty 
into the outcomes of actions \cite{daniele1999strong}. 
Solutions to FOND problems are \emph{policies}, i.e., 
mappings from states to actions, that reach a goal 
state when executed from the initial state of problems. 
The execution of a policy may result in more than one 
sequence of states. Therefore, solutions to FOND 
planning problems are characterized with respect 
to the different possible executions of plans. 
In \cite{cimatti2003weak} three classes of solutions are 
presented: (\emph{i}) \emph{weak solutions} having a chance 
to achieve the goal; (\emph{ii}) \emph{strong solutions} 
guaranteed to achieve the goal; and (\emph{iii}) \emph{strong 
cyclic solutions} having a possibility to terminate and 
if they terminate, guaranteed to achieve~the~goal. 

Strong cyclic solutions have become an important concept in FOND 
planning as they deal with \emph{deadends}, i.e., unsolvable 
states where replanning is not possible. 
As a result, notable FOND planning techniques have been 
developed over the past years including planners relying 
on classical algorithms (referred to as replanners) such 
as PRP \cite{muise2012improved}, FIP \cite{fu2011simple}, 
and NDP \cite{kuter2008using}, and non-classical planners 
relying on binary decision diagrams (BDDs) such as MBP 
\cite{cimatti2003weak}, 
and SAT planning such as FOND-SAT \cite{geffner2018compact}.

Replanners compute a weak policy that does not address all 
non-deterministic reachable states. During the execution 
if an unpredicted state is reached, they compute a new policy 
starting from the unpredicted state. They repeat this process 
until every possible execution of the policy achieves the goal. 
Replanners have been shown to scale up best to 
existing FOND benchmarks from the past International Planning 
Competitions (IPC) \cite{muise2012improved,fu2011simple}. 
However, in \cite{geffner2018compact}, authors show that most 
of the FOND benchmarks from IPC do not pose a big challenge 
to replanners, except for large problems. 
They introduce domains that give rise to an exponential 
number of \emph{misleading plans}, i.e., weak plans involving 
unsolvable states that do not lead to strong policies, however, 
due to their minimal lengths, are likely to be found by replanners, 
and show existing replanners tend to break on these problems. 

Replanners strongly rely on their adopted determinization strategy 
and the initial weak plan that they generate and refine toward a 
final strong solution. 
There are two known methods for compiling from non-deterministic 
to deterministic domains \cite{yoon2007ff}: 
(\emph{i}) \emph{single-outcome} which selects and includes only 
one outcome for each non-deterministic operator in the deterministic 
domain, and (\emph{ii}) \emph{all-outcome} which selects and includes 
every outcome as a distinct operator in the deterministic domain. 
All-outcome replanners compile a non-deterministic domain into 
a single deterministic domain, leading to continually 
generating misleading plans (in the presence of such plans 
in a FOND problem).
Single-outcome replanners, by contrast, compile a non-deterministic 
domain into a set of deterministic domains which allow to compute 
varying weak plans that some of them might directly lead to full 
strong policies. 

We present \SP (SP), an offline replanner motivated by NDP2 
\cite{alford2014plan}, that makes the following contribution: 
SP adopts a single-outcome determinization to~solve~FOND 
problems involving misleading plans. 
SP generates a set of single-outcome deterministic domains 
from a non-deter\-ministic domain and ranks them according to 
an ordering heuristic, e.g., 
on the number of planning operators' effects.
In order not to lose the completeness, SP also includes all-outcome 
determinization in the set of compiled domains.
We show experimentally this determinization strategy allows 
SP to compute weak plans that can directly lead to strong 
policies in problems involving misleading plans.
SP is provably shown to be sound and complete and 
its code is publicly provided at:
\sans{https://github.com/mokhtarivahid/safe-planner}.

SP adopts a similar strategy of NDP2 in solving FOND problems, 
i.e., in dealing with unsolvable states and computing weak plans, 
except for the determinization approach. 
SP integrates off-the-shelf classical planners for its internal 
reasoning without any modification. When deadends are identified 
during FOND planning, 
SP modifies a planning problem to prevent a classical planner 
from finding weak plans involving actions that lead to deadends. 
SP also adopts a \emph{multiprocessing} approach; that is, SP 
simultaneously calls different solvers to make weak plans and 
as soon as one solver finds a plan, it terminates other solvers, 
thus exploiting the strength of each individual planner in 
solving problems, {e.g., heuristic state space planners have 
been considered stronger on many non-optimal planning problems 
and SAT planners usually give optimality guarantees}.

We evaluate the performance of SP in a set of existing FOND 
benchmarks from the past IPCs and literature. We compare 
our results to PRP and FOND-SAT, as~well~as~the basic algorithm 
of NDP2, and 
show that SP outperforms these FOND solvers, particularly in 
the introduced domains in \cite{geffner2018compact} involving
a large number of misleading plans. The experiments show 
the adopted single-outcome determinization can have a remarkable 
impact on replanners' performance. We also present successful 
integrations of SP into simulated and real robot applications 
in non-deterministic tasks.

\section{Related Work}
\label{sec:literature}

FF-Replan \cite{yoon2007ff} is an early replanner that uses a 
classical planner for reasoning. FF-Replan is a reactive online 
planning algorithm that generates all-outcomes determinization 
of a non-deterministic domain and using the classical planner 
FF \cite{hoffmann2001ff} makes a weak policy to a problem.
If the execution of the weak policy fails in an unexpected 
state in the non-deterministic environment, FF is reinvoked 
to make a new weak policy from the failed state. FF-Replan 
repeats this procedure until a goal state is reached. 
FF-Replan has been demonstrated to be very effective in many 
non-deterministic planning problems, 
however, it does not generate strong policies, and is hence 
oblivious to getting stuck in deadends. 

RFF \cite{teichteil2008rff} and NDP \cite{kuter2008using} are 
two offline replanners that iteratively expand an initial 
plan to generate a robust policy to a problem. RFF, the winner 
of the probabilistic track~of the 6th international planning 
competition (IPC2008), 
computes an initial weak policy by solving the all-outcome 
determinization of a problem using FF and then iteratively 
improves the robustness of the policy by finding a failed state 
in the non-deterministic environment and replanning from that 
failed state to a goal state. RFF terminates as soon as the 
probability to reach some failed state from the initial state 
is less than a threshold. RFF does not guarantee to avoid deadends, 
hence the generated policies are not strong solutions.
A similar idea is present in NDP, however, during improving 
the robustness of a policy, NDP ignores any probabilities 
leading to deadends and thus produces strong cyclic solutions.
The same group also presents NDP2 \cite{alford2014plan},
an extended version of NDP with some major improvements.
SP~implements~an NDP2 strong cyclic planning algorithm~that 
relies on a~single-outcome determinization leading to avoid 
generating misleading plans, and thus showing an improved 
performance.

FIP \cite{fu2011simple} is a customized FF-replanner which 
introduces some optimizations into a basic algorithm inspired 
by NDP. When searching for a weak plan, FIP directly removes 
state-action pairs that lead to deadends from the search tree, 
and stops the search when a solved state is revisited. FIP 
also introduces a goal alternative heuristic. That is, to 
recover from a failed state, FIP first attempts to find a 
plan for the intended effect of that failed state before 
replanning to the overall goal of the problem. Despite 
showing a good performance, FIP faces the same limitation 
on dealing with 
misleading plans due to its all-outcome determinization.

PRP \cite{muise2012improved} is an all-outcome replanner to 
compute strong cyclic plans. Instead of storing explicit 
state-action pairs, PRP stores a mapping of partial states 
to actions. A partial state is a regression from the goal to 
an action, containing only the relevant portion of an intended 
complete state. 
PRP detects deadends, by a fast but incomplete verification 
algorithm, in the delete relaxation of all-outcome deterministic 
planning, and computes a minimal partial state for each detected 
deadend state. PRP then, rather than backtracking, resets the 
policy and starts the computation over with the knowledge of 
deadend state-action pairs to restrict the expansion of such nodes. 
PRP shows to scale up best among the existing replanners, however 
it faces similar limitations on dealing with misleading plans.

FOND-SAT \cite{geffner2018compact} introduces a compact 
SAT formulation for FOND planning. It relies on CNF encodings 
of atoms~and non-deterministic actions of 
a FOND problem indexed by controller states, and running 
a SAT-solver over these encodings.
FOND-SAT produces compact policies from a satisfying 
truth assignment of the problem. 
It performs well in problems involving
many misleading plans where the problems are not too large.
However, the challenge for FOND-SAT still remains at scaling 
up to problem~size~and~policy~size.

MBP \cite{cimatti2003weak} is one of the best known strong 
cyclic planners that formulates planning as model-checking. 
MBP encodes a domain description into a BDD-based representation 
which allows a compact representation of large sets of states. MBP 
then implements a wide range of algorithms, e.g., weak, 
strong, strong cyclic, conformant, contingent and partial observability, 
that could operate on top of this representation. The inherent weakness 
of MBP is that it does not exploit any heuristics used as in classical 
planning and has to search a very large state space in most planning 
problems, and hence does not scale up to problems of large sizes.

%% file: 2-formulation.tex
\section{FOND Planning Problems}
\label{sec:formalization}

We present a formalism for modeling FOND planning problems which 
is equivalent to the PDDL representation with ``oneof'' clauses 
\cite{bonet20055th}.

A \emph{non-deterministic planning domain} is described by a pair, 
$D=(L,O)$, where $L$ is a \emph{first-order language} that has a 
finite number of predicate symbols and constant symbols to represent 
various properties of the world and $O$ is a set of non-deterministic 
planning operators (as follows). 

Every \emph{non-deterministic planning operator} $o\in O$ is specified 
by a tuple, $(h,P,E)$, where $h$ is the operator's head, a functional 
expression of the form $n(x_1,\dots,x_k)$ in which $n$ is the name 
and $x_1,\dots,x_k$ are the variables appearing anywhere in $o$, 
$P$ is the precondition of $o$, a set of literals that must be proved 
in a world state in order for $o$ to be applicable in that state, 
and $E=\{e_1,\dots,e_m\}$ is a set of non-deterministic effects
of $o$ such that each $e_i$ is 
a set of literals specifying the changes on a state effected by $o$. 
The effects of $o$ are mutually exclusive, that is, only one 
effect happens at a time: $\ts{Probability}(e_i \land e_j)=0$ 
for all pairs of integers $i,j\in \{1,\dots,m\}$ with $i \neq j$.

The negative and positive preconditions of an operator are
denoted as $P^-$ and $P^+$.
The negative effect, $e_i^-$, and the positive effect, 
$e_i^+$, are similarly defined for every $e_i \in E$. 

A \emph{state} is a set of ground predicates 
of $L$ that represents all properties of the world. 
The set of all possible states of the world is denoted 
by ${S}\subseteq 2^{\{\text{all ground atoms of }L\}}$.

An \emph{action} 
$a=(n(c_1,\dots,c_k),P,E)$ is a ground instance of a non-deterministic 
operator $(n(x_1,\dots,x_k),P,E)\in{O}$, where each $c_i$ is a constant 
symbol of ${L}$ that instantiates a variable $x_i$ in the operator description. 

In any given state $s$, if $P^+ \subset s$ and $P^- \cap s = \emptyset$, 
then $a$ is applicable to $s$ and the result of applying $a$ to $s$ 
is a set of states given by the following state-transition function: 
$\gamma(s,a) = \{(s - e_i^-) \cup e_i^+ \mid e_i \in E,~1\leq i\leq m\}.$

The set of all possible actions, i.e., ground instances 
of planning operators in $O$, is denoted by $A$.

A fully observable \emph{planning problem} $\mc{P}$ for a domain $D$ is 
a tuple, \PP, where $s_0 \in S$ is the initial state, and $g$ is the goal, 
i.e., a set of propositions to be satisfied in a goal state $s_g \in S$.

A \emph{policy} $\pi$ is a set of pairs $(s,a)$, where each $s\in S$ 
is a unique state in $\pi$ and each $a\in A$ is an applicable action in 
$s$, denoted by $\pi(s)=\{a \mid (s,a)\in\pi,~\gamma(s,a)\neq\emptyset\}$. 

Given a policy $\pi$, the set of all states reachable by 
following $\pi$ from $s_0$ is denoted by $S_\pi \subseteq S$:
$S_\pi = \{s \mid (s,a)\in\pi\}$. 

The \emph{execution} of a policy $\pi$ starting from a state $s$, denoted 
by $\Sigma_\pi(s)$, is a function that results in a set of terminal states:
\begin{algorithm}[!ht]
\vspace{-15pt}
\small
\DontPrintSemicolon
\SetAlgoNoEnd
\SetKwProg{Fn}{function}{}{end}
\Fn{$\Sigma_\pi(s)$}{
  \lIf{$\pi=\emptyset$}{\Return $\{s\}$}
  $T\gets\{\},~Q\gets \langle s\rangle,~V\gets \{\}$\\
  \While {$Q\neq \emptyset$}{
    $s\gets Q.\ts{pop}(),~V.\ts{add}(s)$\\
    \For{\upshape$s'\in\gamma(s,\pi(s))\text{ s.t. }s'\notin V$}{
      \lIf{$s'\in S_\pi$}{$Q.\ts{add}(s')$}
      \lElse{$T.\ts{add}(s')$}
    }
  }
  \Return $T$
}
\vspace{-10pt}
\end{algorithm}

A policy $\pi$ is a \emph{solution} for a problem $\mc{P}=(D,s_0,g)$ 
if: $|\Sigma_\pi(s_0)| > 0$, that is, there are no \emph{inescapable} 
cycles in $\pi$ which result in zero terminal states; 
and the execution of $\pi$ eventually terminates in a goal state $s_g$ 
such that $g \subseteq s_g$:

\begin{itemize}
\item
$\pi$ is \emph{weak}, if $\exists s \in \Sigma_\pi(s_0):g \not\subset s$;

\item
$\pi$ is \emph{strong}, if $\forall s\in\Sigma_\pi(s_0):g \subseteq s$.

\end{itemize}

A strong policy $\pi$ is \emph{acyclic} if the execution of $\pi$ 
never visits the same state twice; and \emph{cyclic} otherwise, however 
guaranteed to eventually reach the goal in every fair execution of $\pi$ 
\cite{cimatti2003weak}.

\section{Classical Planning Problems}
\label{sec:classical}

Classical planning domains are simplified models of non-deterministic 
planning domains where the state-transition function $\gamma$ returns only one 
possible state per action: $|\gamma(s,a)|=1,~\forall s\in S,~\forall a\in A$.
Simplifying the notation and definitions given above for FOND problems: 
a planning operator $o=(h,P,E)$ is \emph{classical} if $|E|=1$, i.e., $E$ 
is the single effect of $o$. Thus, the result of applying an action $a$ 
in a state $s$ is a new single state given by $\gamma(s,a) = (s - E^-) \cup E^+$.

Intuitively, a domain $D=(L,O)$ is classical if $O$ is a set of 
classical operators, and a problem $\mc{P}=(D,s_0,g)$
is classical if $D$ is a classical domain.

Solutions to classical planning problems are conventionally 
sequential plans rather than policies. Let $\mc{P}=(D,s_0,g)$ be 
a classical planning problem. A plan $p$ is any sequence of actions 
$\langle a_1,\dots,a_k \rangle$ for $k\geq0$, such that there 
exists a sequence of states $\langle s_0,\dots,s_k\rangle$ 
where each $s_i=\gamma(s_{i-1},a_i)$ for~$1\leq i \leq k$. 
The plan $p$ is a solution to $\mc{P}$ if $g \subseteq s_k$.

\begin{definition}[\textbf{Acyclic Policy Image}]
\label{def:policy_image}
Let $\mc{P}=(D,s_0,g)$ be a classical planning problem and 
$p=\langle a_1,\dots,a_k \rangle$ an \emph{acyclic} plan for 
$\mc{P}$. The plan $p$ can be translated into an equivalent 
policy $\pi=\{(s_0,a_1),\dots,(s_{k-1},a_k)\}$, called the 
\emph{acyclic policy image} of $p$ for $\mc{P}$, such that 
both $p$ and $\pi$ produce exactly the same sequence of state 
transitions at $s_0$, and each $s_i=\gamma(s_{i-1},a_i)$ 
for~$1\leq i \leq k$.
\end{definition}

%% file: 3-safe-planner.tex
\section{Compilation from FOND to Classical}
\label{sec:compilation}

Let $o=(h,P,\allowbreak\{e_1,\dots,e_m\})$ be a non-deterministic 
operator. The classical \emph{compilation} of $o$ is a set of 
classical operators $\{o_1,\dots,o_m\}$ such that each 
$o_i=(h,P,e_i)$ for $1\leq i\leq m$.
We employ a single-outcome compilation to translate from 
non-deterministic to classical, and generate a set of all 
possible classical domains from the combination of the 
non-deterministic planning operators' outcomes. 

Algorithm~\ref{alg:compile} shows our method of compiling 
a non-deterministic domain into a set of classical domains. 
It takes as input a non-deterministic domain $D=(L,O)$ 
including $n$ non-deterministic operators, and compiles it 
into an ordered set of classical planning domains, $\Delta$. 
We first build a set of all possible combinations of 
the non-deterministic planning operators' outcomes 
(line~\ref{alg:compile:prod}). We employ the $n$-ary 
Cartesian product over the $n$ sets of the non-deterministic 
effects of the planning operators in $O$. Then, for every 
ordered tuple of effects, $(e_1,\dots,e_n)$, we create and 
add a new classical domain to $\Delta$ by replacing the effects 
of every non-deterministic planning operator with every $e_i$ 
(lines~\ref{alg:compile:startloop}-\ref{alg:compile:endloop}).

\SetInd{5pt}{8pt}
\RestyleAlgo{ruled}
\SetCommentSty{scriptsize}
\begin{algorithm}[!t]
\small
\caption{Compilation from FOND to Classical\label{alg:compile}}
\DontPrintSemicolon
\LinesNumbered
\SetKwComment{tcp}{$\triangleright$\ }{}

\KwIn{\ND\tcp*[f]{a non-deterministic domain}}
\KwOut{$\Delta$\tcp*[f]{an ordered set of classical domains}}
\BlankLine
$\Delta \gets \{\}$\\
$E^n = E_1\times\dots\times E_n=\{(e_1,\dots,e_n)|e_i\in E_i,1\leq i\leq n\}$\label{alg:compile:prod}\\
\For{\upshape$(e_1,\dots,e_n)\in \text{sorted}(E^n)$}{\label{alg:compile:startloop}
    $\overline{O} \gets \{\}$
    \tcp*[f]{a set of classical operators}\\
    \For(\tcp*[f]{$i\in\{1,\dots,n\}$})
    {$o_i\in O$}{
      $\overline{O}\gets \overline{O} \cup (o_i.h,o_i.P,e_i)$
    }
    $\Delta \gets \Delta \cup (L,\overline{O})$\label{alg:compile:endloop}
}

\Return $\Delta \cup (L,\bigcup_{o\in O} \bigcup_{e\in o.E} (o.h,o.P,e))$
\label{alg:compile:return}

\end{algorithm}

In \cite{vallati2015effective}, it is shown that the performance 
of classical planners is affected by domain model configuration.
In Algorithm~\ref{alg:compile}, the obtained classical domains 
are also ranked according to the quality of non-deterministic 
operators, e.g., 
operators with a larger 
number of (or fewer) effects are given the highest priority.
We note there are problems where this heuristic might not be 
helpful in avoiding misleading plans, however, in practice, 
it works well on most of the benchmark problems. 
In this work, we do not investigate the impact of other 
possible heuristics for reordering planning operators. 
A detailed study of different heuristics has been conducted 
in \cite{vallati2015effective,vallati2017improving}, which 
can be used and integrated in our planner. 

It is worth noting that in practice the single-outcome compilation 
was sufficient to solve existing FOND benchmarks, however, there 
are situations where the single-outcome compilation can lead to 
incompleteness. 
{For example, let us consider a planning task with two variables $x$ 
and $y$, both of which are initially false. The goal is to make two 
variables, $x$ and $y$, true. Furthermore, there is only one action 
$a$ with two possible effects: either $x$ is made true or $y$ is made 
true. There is a strong cyclic solution, namely the application of the 
action $a$ in the initial state, that leads into two states in which 
either $x$ or $y$ is true. The single-outcome yields two planning domains, 
a domain with a single action $\langle\text{true},x\rangle$ and a domain 
with a single action $\langle\text{true},y\rangle$. 
Since there is only one relevant action in each domain, 
both domains lead into unsolvable tasks. 
}
So, in order not to lose the completeness, we also append all-outcome 
determinization at the end of the list of compiled domains
(line~\ref{alg:compile:return}).

\section{Planning for Strong Cyclic Solutions}
\label{sec:planning}

We present a non-deterministic planner, \SP (SP), for computing 
strong cyclic policies to FOND problems. 
SP is a variant implementation of NDP2 \cite{alford2014plan} that 
relies on a single-outcome determinization.
We first describe the main algorithm of SP (Algorithm~\ref{alg:sp}) 
which compiles a planning domain from non-deterministic to classical 
and aggregates solutions to classical problems and forms 
a strong policy. Then, we elaborate on the internal functions 
of SP to avoid deadends and compute acyclic plans 
(Algorithms~\ref{alg:cons}~and~\ref{alg:fp}).

In Algorithm~\ref{alg:sp}, SP takes as input a FOND problem $\mc{P}$ 
including a non-deterministic domain $D$, an~initial state $s_0$, 
and a goal $g$, and a set of external classical planners $X$, 
and as output computes a strong~(possibly cyclic) policy $\pi$ 
for~$\mc{P}$.

On line~\ref{alg:sp:comp}, SP compiles $D$ into a set of 
classical domains $\Delta$ (see Algorithm~\ref{alg:compile}). 
On line~\ref{alg:sp:deadend}, $\pi$ is the current evolved 
strong policy, and $\theta$ retains a set of deadend 
(unsolvable)~states. 

On each planning iteration (lines~\ref{alg:sp:while}-\ref{alg:sp:rm}), 
on line~\ref{alg:sp:while}, SP finds a non-goal terminal state $s$ by 
the execution of the current policy $\pi$ on the initial state: 
$\Sigma_{\pi}(s_0)$. Note that when $\pi$ is empty (e.g., 
in the first iteration) $s$ is the initial state.

\SetInd{5pt}{8pt}
\RestyleAlgo{ruled}
\begin{algorithm}[!t]
\small
\caption{\SP (SP)\label{alg:sp}}
\DontPrintSemicolon
\SetKwComment{tcp}{$\triangleright$\ }{}

\KwIn{(\PP,~$X$)
  \tcp*[f]{$X$ is a set of classical planners}
}
\KwOut{
  $\pi$\tcp*{a strong cyclic solution}
}
\BlankLine
$\Delta \gets \text{Compilation}(D)$\label{alg:sp:comp}
\tcp*[f]{see Alg.~\ref{alg:compile}}\\
$\pi \gets \{\},~\theta \gets \{\}$\label{alg:sp:deadend}
\tcp*[f]{a set of deadend states}\\
\While
{\upshape$\exists s\in \Sigma_{\pi}(s_0) \text{ s.t. } g\not\subset s$\label{alg:sp:while}}{
  $\pi' \gets \text{\text{\MSP}}(X,D,\Delta,s,g,\theta)$\label{alg:sp:find}
  \tcp*[f]{see Alg.~\ref{alg:fp}}\\
  \If{\upshape $\pi' \neq \text{failure}$\label{alg:sp:if}}{
    $\pi \gets \pi \cup \{(s',a)\mid(s',a') \in \pi',\text{ $a$ in $D$ corresp. $a'$}\}$\label{alg:sp:merge}
  }
  \Else{\label{alg:sp:else}
    \lIf(\tcp*[f]{no plan exists at $s_0$}){$s=s_0$\label{alg:sp:elif}}{
        \Return failure
    }
    $\theta \gets \theta \cup \{s\}$\label{alg:sp:deadend_s}
    \tcp*[f]{$s$ is a deadend state}\\
    \For
    {\upshape$(s',a')\in\pi \text{~s.t.~} s\in \gamma_D(s',a')$\label{alg:sp:deadend_loop}}{
      $\pi\gets \pi \setminus \{(s',a')\}$\label{alg:sp:rm}
      \tcp*[f]{remove $(s',a')$ from $\pi$}\\
    }
  }
}
\Return $\pi$
\label{alg:sp:solution}

\end{algorithm}

On line~\ref{alg:sp:find}, \MSP (explained later in 
Algorithm~\ref{alg:fp}) is a function that given the 
tuple $(X,D,\Delta,s,g,\theta)$ computes an acyclic 
policy image $\pi'$ at $s$ (if any) which never produces 
states in $\theta$ nor visits the same state~twice.

If there is such a policy image $\pi'$ at $s$ (line~\ref{alg:sp:if}), 
then it is merged into $\pi$ with forcing to replace the old states 
in $\pi$ with new ones in $\pi'$ (line~\ref{alg:sp:merge}), that is, 
if there is a common state $s'$ in $\pi$ and $\pi'$, old actions 
in $\pi(s')$ are replaced by new actions in $\pi'(s')$.
This is the key to avoiding inescapable loops in $\pi$ whereby 
neither a goal nor a non-goal terminal state could be found on 
the next iterations of SP. 
Note that when merging $\pi'$ into $\pi$, the deterministic actions 
in $\pi'$ are replaced with their corresponding non-deterministic 
actions in $D$ (line~\ref{alg:sp:merge}).

If no such policy image exists at $s$ (line~\ref{alg:sp:else}), 
and $s$ is the initial state (line~\ref{alg:sp:elif}), SP returns 
a failure; otherwise SP concludes that $s$ is a deadend state 
and adds it to $\theta$ (line~\ref{alg:sp:deadend_s}). 
SP also removes from $\pi$ all states and actions in $\pi$ 
that produce $s$ (lines~\ref{alg:sp:deadend_loop}-\ref{alg:sp:rm}). 

SP repeats this procedure until a strong cyclic solution is computed
(lines~\ref{alg:sp:solution}) and there are no more non-goal terminal 
states in $\Sigma_{\pi}(s_0)$ (line~\ref{alg:sp:while}); or all deadend 
states are detected and removed from $\pi$ and no more plan exists at $s_0$ that 
avoids deadends, where planning ends in failure (line~\ref{alg:sp:elif}).

\subsection{Constraining a Domain and a Problem}
\label{sec:constrain}

The core of SP is to avoid deadends (and cycles) when computing an 
acyclic policy image at a non-goal terminal state 
(shown later in Algorithm~\ref{alg:fp}). SP uses classical 
planners as a black box and does not modify the internal algorithms 
of the planners. Therefore SP must deal with deadends externally.
An approach to avoiding deadends is to look~for alternative plans 
that do not start with already known actions leading to deadends.
A method for modifying a classical planning domain and a planning 
problem is proposed in \cite{alford2014plan} such that actions 
leading to deadends become inapplicable at the first step of any 
solution to the problem.
Based on that work, we 
present the Constrain function in Algorithm~\ref{alg:cons}.
It receives as input a classical domain $D$, a state 
$s$ and a set of uninteresting actions $A$ that lead to deadends
(or cycles)
when applied in $s$, and returns a constrained domain $D'$ and 
a constrained state $s'$ that ensure a classical planner cannot 
produce a plan starting with an action in $A$ (if any).

\RestyleAlgo{ruled}
\begin{algorithm}[!t]
\small
\caption{Constrain a Domain and a Problem\label{alg:cons}}
\DontPrintSemicolon
\SetKwComment{tcp}{$\triangleright$\ }{}

\KwIn{$(D,s,A)$
\tcp*[f]{$A$ is a set of actions leading to deadends at $s$}\\
}
\KwOut{$(D',s')$\tcp*[f]{(a constrained domain, a constrained state)}}
\BlankLine
$D' \gets D$,
$s' \gets s$,
$(L',O') \gets D'$\label{alg:cons:copy}\\
\ForEach{\upshape action $n(c_1,\dots,c_k)\in A$}{\label{alg:cons:s1}
    $s' \gets s' \cup \{\text{\sans{disallowed}}_n(c_1,\dots,c_k)\}$\label{alg:cons:s2}\\
    $L' \gets L' \cup \{\text{\sans{disallowed}}_n(x_1,\dots,x_k)\}$\label{alg:cons:l}\\
}

\ForEach{\upshape operator $(n(x_1,\dots,x_k),P,E)\in O'$\label{alg:cons:o}}{
    \ForEach{\upshape action $n(c_1,\dots,c_k)\in A$\label{alg:cons:p}}{
        $P \gets P \cup \{\neg\text{\sans{disallowed}}_n(x_1,\dots,x_k)\}$\label{alg:cons:p2}\\
    }
    \ForEach{\upshape action $m(c_1,\dots,c_l)\in A$\label{alg:cons:e}}{
        $E \gets E \cup \{\neg\text{\sans{disallowed}}_m(c_1,\dots,c_l)\}$\label{alg:cons:e2}\\
    }
}

\Return $(D',s')$
\label{alg:cons:return}
\end{algorithm}

On line~\ref{alg:cons:copy}, $D$ and $s$ are preserved unmodified 
by making a copy of them into $D'$ and $s'$.
For every action $n(c_1,\dots,c_k)$ in $A$, a ground predicate 
$\text{\sans{disallowed}}_n(c_1,\dots,c_k)$ is introduced into $s'$ 
(lines~\ref{alg:cons:s1}~and~\ref{alg:cons:s2}), and an ungrounded 
predicate $\text{\sans{disallowed}}_n(x_1,\dots,x_k)$ is introduced 
into $L'$ (line~\ref{alg:cons:l}).

For every planning operator $(n(x_1,\dots,x_k),P,E)\in O'$
(line~\ref{alg:cons:o}) that has instances in $A$ 
(line~\ref{alg:cons:p}), a negated ungrounded predicate 
$\neg\text{\sans{disallowed}}_n(x_1,\dots,x_k)$ is introduced 
into the operator's precondition $P$ (line~\ref{alg:cons:p2}), 
and for all actions $m(c_1,\dots,c_l)\in A$ (line~\ref{alg:cons:e}), 
a negated ground predicate 
$\neg\text{\sans{disallowed}}_m(c_1,\dots,c_l)$ is introduced 
into the operator's effect $E$ (line~\ref{alg:cons:e2}).

This ensures a grounding of an operator $n(x_1,\dots,x_k)$ for 
constants $(c_1,\dots,c_k)$ never occurs in $s'$ containing the 
predicate $\text{\sans{disallowed}}_n(c_1,\dots,c_k)$, as well as all 
predicates $\text{\sans{disallowed}}_m(c_1,\dots,c_l)$ are immediately 
removed from $s'$ when the first action in $D'$ is applied in $s'$.

\subsection{Computing an Acyclic Policy Image}
\label{sec:fp}

In Algorithm~\ref{alg:fp}, 
\MSP (MSP) is a function that 
given a tuple of a set of classical planners $X$, a~non-deter\-ministic 
domain $D$, a set of classical domains $\Delta$, an initial state $s_0$, 
a goal $g$, and a set of deadend states $\theta$, and using the above 
Constrain function, computes an acyclic policy image $\pi$ at $s_0$ (if 
any) that never produces states in $\theta$ nor~has~cycles.

On line~\ref{alg:fp:init}, $\pi$ is the current acyclic policy 
image, $B$ is a set of state-action pairs leading to cycles 
or deadends which no plan can begin with, $U$ is a set 
of deadend states which cannot take part in any solution, 
and $s$ is the current state. 

Within an infinite loop (line~\ref{alg:fp:loop}), MSP iterates 
over the compiled classical domains in $\Delta$ 
(lines~\ref{alg:fp:outfor}-\ref{alg:fp:addB2}). On each iteration, 
on line~\ref{alg:fp:cons}, MSP constrains $\overline{D}$ and $s$ into 
$\overline{D'}$ and $s'$ such that no plan can be found that begins 
with actions associated with $s$ in $B$ (see Algorithm~\ref{alg:cons}).
On line~\ref{alg:fp:if1}, Planner is a \emph{multiprocessing} function 
that calls simultaneously all external planners in $X$ to find 
a plan given the constrained domain $\overline{D'}$ and the 
constrained state $s'$. 
As soon as a plan $\langle \overline{a'}_1,\dots,\overline{a'}_k \rangle$ 
is found by a classical planner, Planner terminates other 
classical planners 
and returns the plan (line~\ref{alg:fp:if1}), and then MSP 
proceeds to merge the whole or part of the plan into $\pi$ 
that avoids producing deadend states in $U$ or causing cycles 
in $\pi$ (lines~\ref{alg:fp:infor}-\ref{alg:fp:update}):

\SetInd{5pt}{8pt}
\RestyleAlgo{ruled}
\begin{algorithm}[!t]
\small
\caption{\MSP (MSP)\label{alg:fp}}
\DontPrintSemicolon
\SetKwComment{tcp}{$\triangleright$\ }{}
\SetKwBlock{Loop}{loop}{EndLoop}

\KwIn{$(X,D,\Delta,s_0,g,\theta)$}
\KwOut{$\pi$\tcp*[f]{an acyclic policy image}}
\BlankLine
  $\pi \gets \langle\rangle;~B \gets \{\};~U \gets \theta;~s \gets s_0$\label{alg:fp:init}\\
  \Loop{\label{alg:fp:loop}
    \For{$\overline{D} \in \Delta$\label{alg:fp:outfor}}{
        $\overline{D'}, s' \gets \text{Constrain}(\overline{D},s,\{a | (s,a) \in B\})$\label{alg:fp:cons}
        \tcp*[f]{Alg.~\ref{alg:cons}}\\
        \If
        {\upshape $\exists \langle \overline{a'}_1,\dots,\overline{a'}_k \rangle 
        \text{ from }\text{Planner}(X,\overline{D'},s',g)$\label{alg:fp:if1}}{
            \For{\upshape $i=1,\dots,k$\label{alg:fp:infor}}{
                $\overline{a}_i \gets \text{the action in $\overline{D}$ corresp. }\overline{a'}_i$\label{alg:fp:corresp1}\\
                $a_i \gets \text{the action in $D$ corresp. }\overline{a}_i$\label{alg:fp:corresp2}\\
                \If{\upshape$\gamma_{\overline{D}}(s,\overline{a}_i) \in S_\pi \lor \gamma_D(s,a_i) \cap U \neq \emptyset$\label{alg:fp:if2}}{
                  $B \gets B \cup \{(s,\overline{a}_i)\}$\label{alg:fp:addB1}\\
                  $\textbf{break}$\label{alg:fp:break1}
                }
                $\pi \gets \pi \cdot (s,\overline{a}_i)$\label{alg:fp:append}\\
                $s \gets \gamma_{\overline{D}}(s,\overline{a}_i)$\label{alg:fp:update}\\
            }
            \Else(\tcp*[f]{inner for-loop finished with no break}){
            \label{alg:fp:else1}
              \Return $\pi$\label{alg:fp:return}
            }
            $\textbf{break}$\label{alg:fp:break2}
            \tcp*[f]{inner for-loop was broken, so break the outer too}\\
        }
    }
    \Else(\tcp*[f]{outer for-loop finished with no solution found for all domains}){\label{alg:fp:else}
      \lIf{\upshape$\pi=\langle\rangle$}{
        \Return failure\label{alg:fp:failure}
      }
      $U \gets U \cup \{s\}$\label{alg:fp:addU}\\
      $(s,a) \gets \pi.pop()$\label{alg:fp:retrieve}\\
      $B \gets B \cup \{(s,a)\}$\label{alg:fp:addB2}\\
    }
  }{}
\end{algorithm}

Lines~\ref{alg:fp:infor}-\ref{alg:fp:break1}:
for each action $\overline{a'}_i$, for $1\leq i\leq k$ 
(line~\ref{alg:fp:infor}), $\overline{a}_i$ is the deterministic 
action in $\overline{D}$ that corresponds to $\overline{a'}_i$ 
(line~\ref{alg:fp:corresp1}) and $a_i$ is the non-deterministic 
action in $D$ that corresponds to $\overline{a}_i$ 
(line~\ref{alg:fp:corresp2}).
On line~\ref{alg:fp:if2}, if the new state produced by $s$ 
and $\overline{a}_i$: $\gamma_{\overline{D}}(s,\overline{a}_i)$, 
is in $S_\pi$ (causing a cycle), or the states produced by 
$s$ and $a_i$: $\gamma_D(s,a_i)$, are in $U$ (leading to 
deadends); then MSP stops merging the plan 
into $\pi$, adds the current state and action $(s,\overline{a}_i)$ 
in $B$ (line~\ref{alg:fp:addB1}), and breaks both for-loops 
(lines~\ref{alg:fp:break1} and \ref{alg:fp:break2}) and starts 
over the planning from $s$ on line~\ref{alg:fp:outfor}.

Lines~\ref{alg:fp:append}-\ref{alg:fp:update}: if $\overline{a}_i$ 
does not cause a cycle and $a_i$ does not lead into deadends, MSP 
appends $(s,\overline{a}_i)$ to $\pi$ (line~\ref{alg:fp:append}), 
and replaces $s$ with the new state (line~\ref{alg:fp:update}), 
and continues testing and merging other actions into $\pi$
(lines~\ref{alg:fp:infor}-\ref{alg:fp:update}).

On line~\ref{alg:fp:return}, if all actions of the current plan are 
merged into $\pi$, that is, no action leads into deadends nor causes 
cycles, MSP terminates and returns the acyclic policy image~$\pi$.

On line~\ref{alg:fp:else}, if no plan exists at $s$ for all 
classical domains in $\Delta$, MSP backtracks from $s$ to a 
previous state and attempts to make a new plan while avoiding $s$: 
MSP adds $s$ to $U$ as an unsolvable state (line~\ref{alg:fp:addU}), 
and retrieves the previous state and action leading to $s$ and 
adds them to $B$ (lines~\ref{alg:fp:retrieve}-\ref{alg:fp:addB2}). 
On line~\ref{alg:fp:failure}, if $\pi$ is empty and there is no 
previous state, then this means there is no plan at $s_0$ that 
can avoid cycles nor states in $U$, so MSP returns failure.

Note lines \ref{alg:fp:else1} and \ref{alg:fp:else} 
conform to the \sans{for-else} Python syntax, meaning 
the associated for-loops finished without~break.

\subsection{Soundness and Completeness}
\label{sec:soundness}

\begin{lemma}
\label{lem:msp}
If planners in $X$ are sound and guaranteed to terminate, then 
MSP is sound and returns either an acyclic policy image that never 
leads to deadends or a failure. 
\end{lemma}

\begin{proof}
MSP returns an acyclic policy image (Alg.~\ref{alg:fp}, 
line~\ref{alg:fp:return}) if it passes the test for all actions 
of the current plan against deadends and cycles, and since the 
planners in $X$ are sound, the last action achieves the goal.
Now, it is enough to show that the current plan never causes 
cycles nor leads to deadends.
By induction, if the policy is empty, then the lemma is 
trivially true. We assume $\langle a_1,\dots,a_{i-1} \rangle$ 
is part of the plan, already processed and merged into $\pi$, 
that avoids states in $U$ and does not cause any cycle 
(lines~\ref{alg:fp:infor}-\ref{alg:fp:update}).
Further assume MSP is adding a new action $a_i$ to $\pi$. 
On line~\ref{alg:fp:if2}, MSP already checks that $a_i$ does 
not lead into deadends nor cause a cycle, and if not, $a_i$ is 
added~in,~so~$\langle a_1,\dots,a_i \rangle$~is~safe. 
\end{proof}

\begin{lemma}
\label{lem:inescapable}
If planners in $X$ are sound, then after every iteration of SP
there are no inescapable cycles in $\pi$.
\end{lemma}

\begin{proof}
Proof by induction. 
By Lemma~\ref{lem:msp}, the returned policy image by MSP, to merge in 
$\pi$, is acyclic, so when $\pi$ is empty, the lemma is trivially true.
For the induction step,~(\emph{i})~MSP returns an acyclic 
policy image, $\pi'$, so SP merges and replaces old states in 
$\pi$ with new ones in $\pi'$ until the end of $\pi'$, which 
by Lemma~\ref{lem:msp} is a goal state. So, any modified states 
in $\pi$ must have a path to a goal; 
(\emph{ii}) MSP returns a failure on a state $s$, so all states 
and actions leading to $s$ are removed from $\pi$, and since 
$s$ was the last non-goal terminal state, any state leading to 
$s$ now becomes a non-goal terminal~state. 
\end{proof}

\begin{theorem}
If planners in $X$ are sound, then SP is sound.
\end{theorem}

\begin{proof}
By Lemma~\ref{lem:inescapable}, after every iteration of SP, 
there are no inescapable cycles in $\pi$. That is, all terminal 
states in $\pi$ are reachable from $s_0$. If SP returns $\pi$, 
which means SP terminates without a failure, then there are no 
more non-goal terminal states in $\pi$, so all states in $\pi$ 
have a path to a goal state, and thus $\pi$ is a valid strong 
policy.
\end{proof}

\begin{theorem}
If planners in $X$ are sound and complete, then SP is complete.
\end{theorem}

\begin{proof}
Proof by contradiction. Let us assume SP is not complete 
and there are a problem $\mc{P}$ and a strong policy $\pi$ 
for $\mc{P}$, such that SP returns a failure to $\mc{P}$.
This means MSP cannot find a plan at $s_0$ that avoids states 
in $\theta$, so $\pi$ has paths from $s_0$ to the goal which 
involve some states in $\theta$. This is a contradiction with 
Lemma~\ref{lem:msp} that MSP never returns a plan including 
deadend states in $\theta$.
\end{proof}

%% file: 4-experiment.tex
\section{Implementation and Empirical Evaluation}
\label{sec:experiments}

\SP has been implemented in {Python} and integrates 
several off-the-shelf classical planners. 
\footnote{\fontsize{8}{8}\selectfont
The code is at:
\sans{https://github.com/mokhtarivahid/safe-planner}}
We present the performance of SP in existing FOND benchmarks 
and validate its utility in real and simulated robotic tasks.

\subsection{Standard FOND Benchmarks}
\label{sec:exp:benchmarks}

We show the performance of SP, using Fast-Downward (FD) 
\cite{helmert2006fast} as its internal planner,
in FOND benchmarks from the past uncertainty tracks 
of the 5th (IPC2006) and the 6th (IPC2008) international 
planning competitions \cite{bonet20055th,bryce20086th}, 
and existing FOND domains from literature 
\cite{muise2012improved,geffner2018compact}. 
We performed all experiments on a machine 1.9GHz Intel 
Core i7 with 16GB memory with the limited planning runtime 
to 30 minutes. We compare our results to PRP 
\footnote{\fontsize{8}{8}\selectfont\sans{https://github.com/qumulab/planner-for-relevant-policies}}
and FOND-SAT 
\footnote{\fontsize{8}{8}\selectfont\sans{https://github.com/tomsons22/FOND-SAT}} 
with their option for computing strong cyclic policies.
Unfortunately, NDP2 is not publicly available, however, 
by disabling the option for the single-outcome determinization 
in SP, we can achieve the basic algorithm of NDP2. So, 
we also report the performance of SP as NDP2 using the 
same planner FD.
Furthermore, we report the results obtained by other 
classical planners integrated into SP, e.g., Fast-Forward 
(FF) \cite{hoffmann2001ff} and Madagascar (M) \cite{RINTANEN201245}.

As mentioned earlier in Algorithm~\ref{alg:compile}, SP ranks 
the compiled single-outcome domains according to the number 
of effects of operators. We ran SP with the best ordering 
heuristic, selected on a domain-by-domain basis by alternating 
between descending (the default ordering heuristic in SP) and 
ascending orders. In Table~\ref{tbl:result}, we report these 
heuristics as $\downarrow$ for descending and $\uparrow$ for 
ascending orders.




\newcolumntype{L}[1]{>{\raggedright\let\newline\\\arraybackslash\hspace{0pt}}m{#1}}
\newcolumntype{C}[1]{>{\centering\let\newline\\\arraybackslash\hspace{0pt}}m{#1}}
\newcolumntype{R}[1]{>{\raggedleft\let\newline\\\arraybackslash\hspace{0pt}}m{#1}}

\begin{table}[t!]
\centering
\resizebox{\linewidth}{!}
{
\begin{tabular}{|l|c|c|c|c|}
\hline
\multirow{1}{*}{Domain (\#problems)} & SP$_{\text{FD}}$ & NDP2$_{\text{FD}}$ & PRP & FOND-SAT \\
\hline
\hline
acrobatics (8)        & \textbf{8} (0)   $\downarrow$ & 6 (0)            & \textbf{8} (0)   & 0 (0)           \\ \hline
beam-walk (11)        & 8 (0)            $\downarrow$ & 8 (0)            & \textbf{11} (0)  & 0 (0)           \\ \hline
blocksworld (30)      & 20 (0)           $\downarrow$ & 20 (0)           & \textbf{30} (0)  & 0 (0)           \\ \hline
doors (15)            & 10 (0)           $\uparrow$   & 10 (0)           & \textbf{12} (3)  & 11 (0)          \\ \hline
elevators (15)        & \textbf{15} (0)  $\downarrow$ & \textbf{15} (0)  & \textbf{15} (0)  & 5 (0)           \\ \hline
ex-blocksworld (15)   & \textbf{6} (6)   $\downarrow$ & \textbf{6} (6)   & \textbf{6} (9)   & 5 (0)           \\ \hline
faults (49)           & 49 (0)           $\uparrow$   & 49 (0)           & \textbf{55} (0)  & 0 (0)           \\ \hline
first-responders (100)& 58 (25)          $\downarrow$ & 58 (25)          & \textbf{75} (25) & 0 (0)           \\ \hline
tireworld (15)        & \textbf{12} (3)  $\downarrow$ & \textbf{12} (3)  & \textbf{12} (3)  & \textbf{12} (0) \\ \hline
triangle-tire (40)    & 3 (0)            $\downarrow$ & 3 (0)            & \textbf{40} (0)  & 2 (0)           \\ \hline
zenotravel (15)       & \textbf{15} (0)  $\downarrow$ & \textbf{15} (0)  & \textbf{15} (0)  & 5 (0)           \\ \hline\hline
islands (60)          & \textbf{60} (0)  $\uparrow$   & 9 (0)            & 31 (0)           & \textbf{60} (0) \\ \hline
miner (50)            & \textbf{50} (0)  $\uparrow$   & 16 (0)           & 14 (0)           & 44 (0)          \\ \hline
tireworld-spiky (11)  & \textbf{11} (0)  $\downarrow$ & \textbf{11} (0)  & 1 (0)            & 9  (0)          \\ \hline
tireworld-truck (74)  & \textbf{73} (0)  $\downarrow$ & 23 (0)           & 21 (0)           & 67 (0)          \\ \hline
\hline
total (514) & \textbf{398} (34) & 289 (34) & 346 (60) & 220 (0) \\ \hline
\end{tabular}
}
\caption{Total number of problems for which SP, NDP2, PRP and 
FOND-SAT compute strong cyclic solutions. The numbers in brackets 
are those problems for which it is proven that no strong solutions 
exist within the 30 minutes. The arrows $\downarrow$ and $\uparrow$ 
indicate respectively the descending and ascending orderings used 
for ranking the compiled single-outcome domains in SP. Best coverages 
are shown in bold.}
\label{tbl:result}
\end{table}

Table~\ref{tbl:result} shows the problem solving coverage for 
SP, NDP2, PRP and FOND-SAT in the used FOND domains. The best 
coverage is shown in bold. 
The domains are shown in two fragments. For the domains in the 
upper part, which are mostly from the past IPCs and do not involve 
many misleading plans, PRP showed an excellent coverage, 
SP and NDP2 showed fairly similar performance, and FOND-SAT 
showed a poor performance.
The last 4 FOND domains, introduced in \cite{geffner2018compact}
where FOND-SAT shows fairly good results, involve many misleading 
plans. Due to the used single-outcome compilation, SP could avoid 
producing misleading plans, 
hence showing better results. 
PRP and NDP2, on the other hand, showed a poor performance in 
these domains because of using only all-outcome determinization.

For an illustration of the relative performance of different planners, 
the cactus diagram in Figure~\ref{fig:cactus} shows the number of 
problem instances that planners solved as the timeout limit is increased. 
Here, we also ran SP with other integrated classical planners, 
e.g., FF and FF+M (multiprocessing).
In most of the domains, SP could achieve the best performance using 
either FD or FF, however, in~some domains, e.g., faults and first-responders,
SP showed better results in multiprocessing mode. 
We note that the~perfor\-mance of SP is degraded slightly using 
multiple planners due to the multiprocessing overhead, e.g., in 
beam-walk, SP showed slightly poor performance~in~multiprocessing. 

\subsection{Real World Tasks} 
\label{sec:exp:robot}

We also evaluated and integrated SP into real world non-deterministic tasks. 
Videos of these tasks are also provided in the supplementary material.

\input{4-1-experiment}

\section{Conclusion}\label{sec:conclusion}

We presented SP, a powerful replanner that relies on a single-outcome 
determinization. Using an ordering heuristic on single-outcome domains, 
SP produced weak plans that could directly lead onto full strong policies. 
SP adopts the basic ideas of NDP2 for FOND planning 
without any state abstraction technique. 
We think the integration of the single-outcome determinization 
with more advanced techniques such as the state relevance in 
PRP can result in considerable improvements in FOND planning.

The future work includes adopting different methods for 
configuring and reordering single-outcome domains, e.g., 
one similar to \cite{vallati2018icaps18}.
Other improvement includes extending SP to solve Partially 
Observable Non-Deterministic (POND) problems and problems 
with multiple initial states.

%% file: 4-1-experiment.tex
\subsubsection{Real Robot Object Manipulation}
\label{sec:exp:real}

We set up a \sans{packaging} task including an automated workspace 
with a dualarm YuMi manipulator and a table containing a set of objects 
that need to be put upward into a box. The orientation of objects on the 
table are not known to the robot. A non-deterministic perception action, 
\sans{check\_orientation}, identifies the orientation of each object as 
either upward or downward.
Given a problem, SP computes a strong policy that gives the robot a 
choice of rotating an object or not at runtime. 
Snapshots of a YuMi manipulator putting two objects into a box
were shown in Figure~\ref{fig:exp:packaging}. 
In this experiment, we used OPTIC \cite{benton2012temporal}, 
a temporal planner integrated into SP, to generate a partially 
ordered strong policy to this task. 
We implemented and executed robot's skills using eTaSL \cite{aertbelien2014etasl}.

\begin{figure}[!t]
  \centering
  \includegraphics[width=\linewidth]{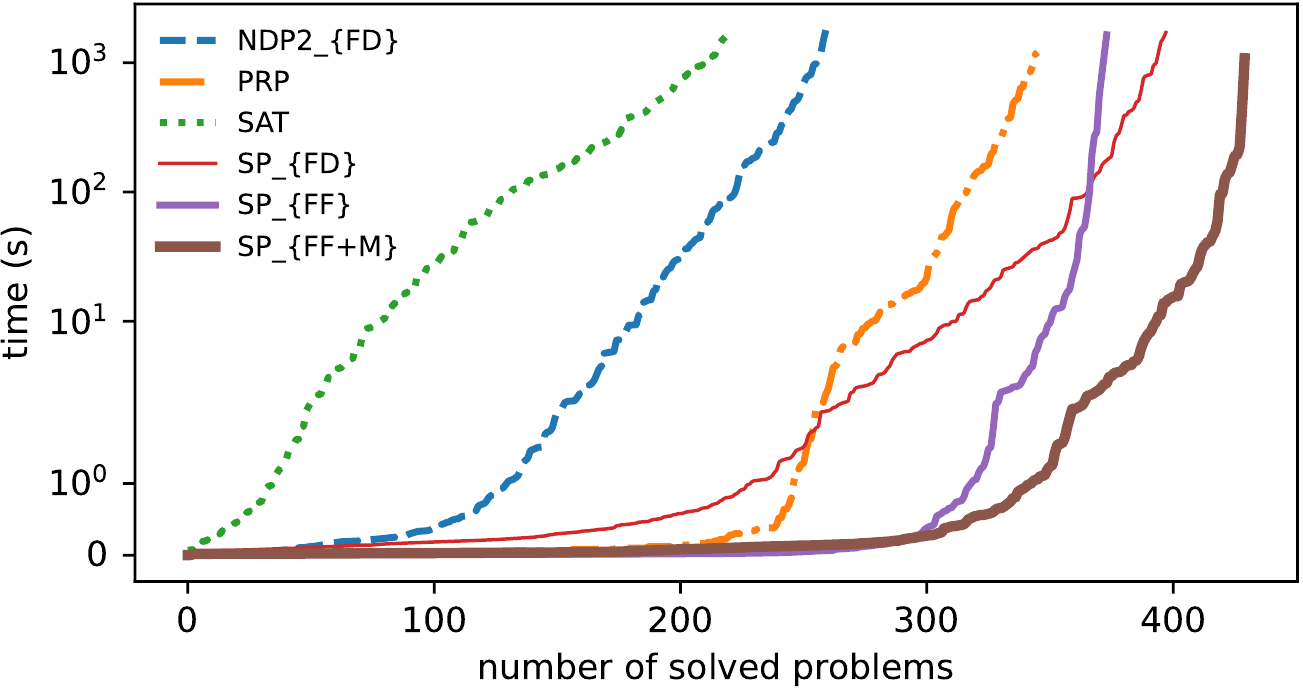}
  \caption{Total number of solved problems as the time is increased.}
  \label{fig:cactus}
\end{figure}

\subsubsection{Simulated Environments}
\label{sec:exp:sim}

We integrated SP into a Task and Motion Planning (TAMP) framework 
\cite{sathya2020task} and demonstrated this framework in two 
simulated non-deterministic tasks. 
We developed a planner-independent interface which translates 
between the high-level symbolic task descriptions and the low-level 
geometrical space. 
In this work, task planning is achieved using SP and motion 
planning is achieved by solving an
Optimal Control Problem (OCP) 
\cite{sathya2020real}. 
More details of this work are given in \cite{sathya2020task}.

\paragraph{Singlearm Tabletop Object Manipulation}
\label{sec:tabletop2}

\begin{figure}[!t]
\begin{center}
  \centering
  \begin{subfigure}[b]{.499\linewidth}
    \includegraphics[width=\textwidth]{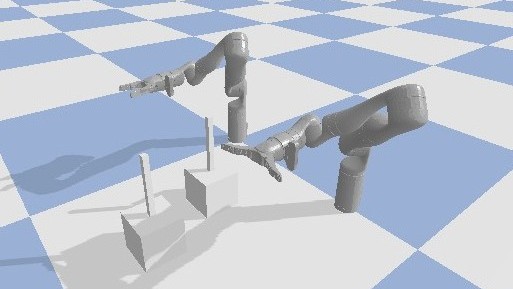}
    \vspace{-17pt}
    \caption{}
  \end{subfigure}\hspace{-3pt}
  \begin{subfigure}[b]{.499\linewidth}
    \includegraphics[width=\textwidth]{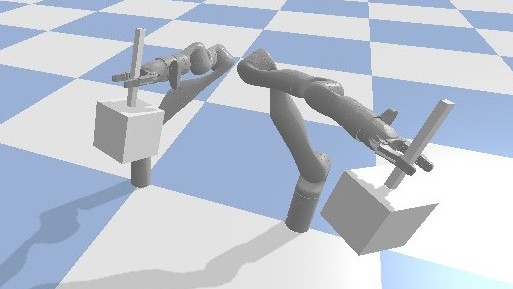}
    \vspace{-17pt}
    \caption{}
  \end{subfigure}
\caption{
The task is to lift 
the farthest object. The object might be heavy,~thus 
the manipulator has to pull it to improve the chance 
of a successful lift, however, the obstructing object 
has to be taken away first.
(a) the initial state; (b) the final configuration 
of the system after executing~the~final~strong~policy.
}
\label{fig:tabletop2}
\end{center}
\end{figure}

\begin{figure}[!t]
  \centering
  \includegraphics[width=\linewidth]{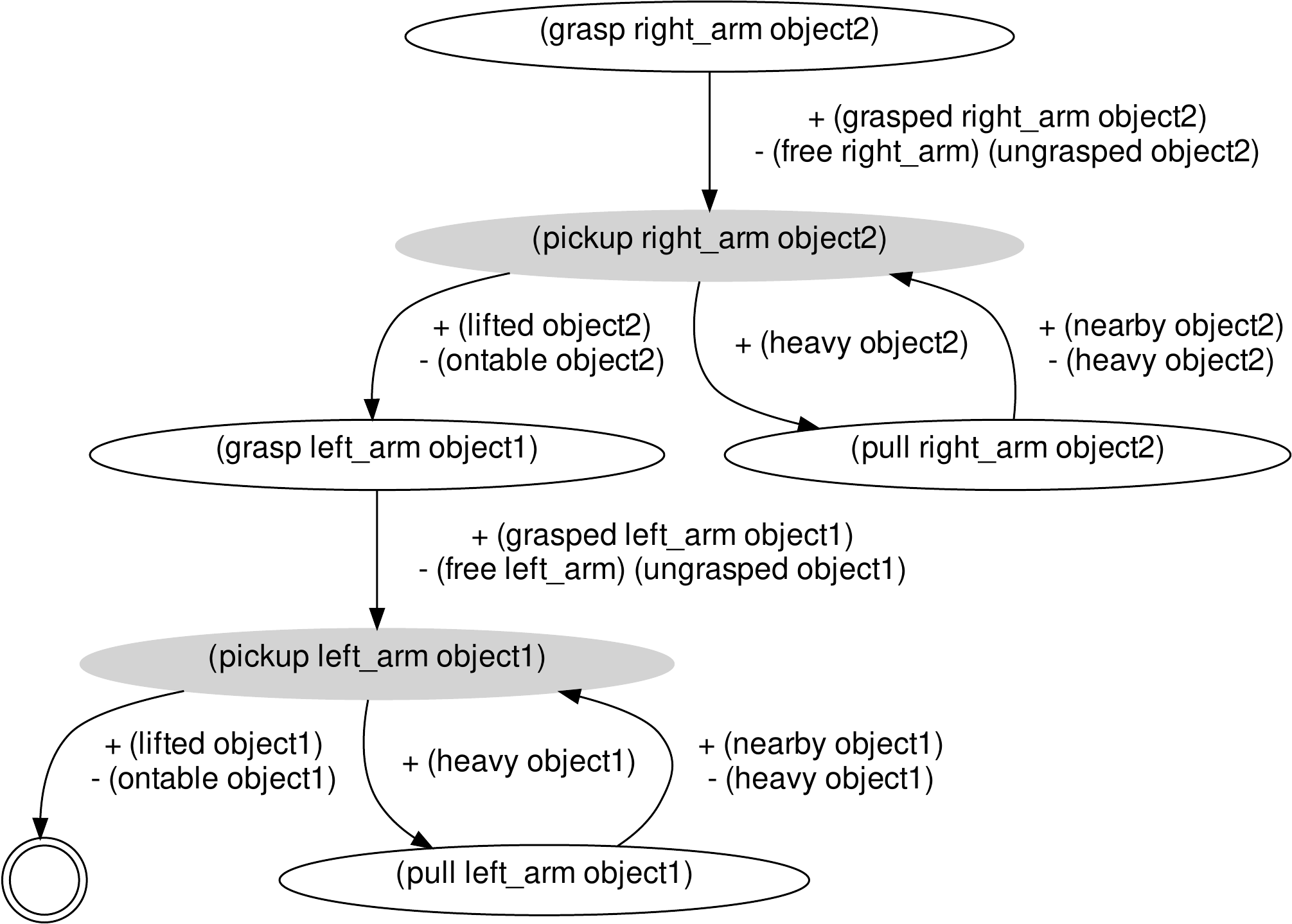}
  \caption{A strong cyclic policy to lift an object, computed by SP.
  Despite involving loops, it is presumed that the plan eventually 
  reaches a goal state under a \emph{fairness} assumption, that is, 
  execution cannot loop forever and outcomes leading to the goal 
  must happen eventually.
  }
  \label{fig:plan_pullpick}
\end{figure}

The first task involves lifting an object where the 
weight of the object is not known to the robot 
(see Figure~\ref{fig:tabletop2}). 
When the object is heavy the lift operation will fail due 
to torque limits, so the robot must first pull the object 
to a configuration that improves the robot's chance of a 
successful lift. We also consider some object that might 
obstruct the path of the pulling object. Therefore, to achieve 
the task, manipulators must first remove obstructing 
objects and then pull and lift the target object.
In this experiment, the TAMP framework tests the feasibility 
of 11 actions, identifies 2 unfeasible actions, and makes 2 
replannings in about 25 seconds.
Figure~\ref{fig:plan_pullpick} also shows the final strong 
cyclic solution, computed by SP, to achieve this task.

\begin{figure}[!t]
\begin{center}
  \centering
  \begin{subfigure}[b]{.497\linewidth}
    \includegraphics[width=\textwidth]{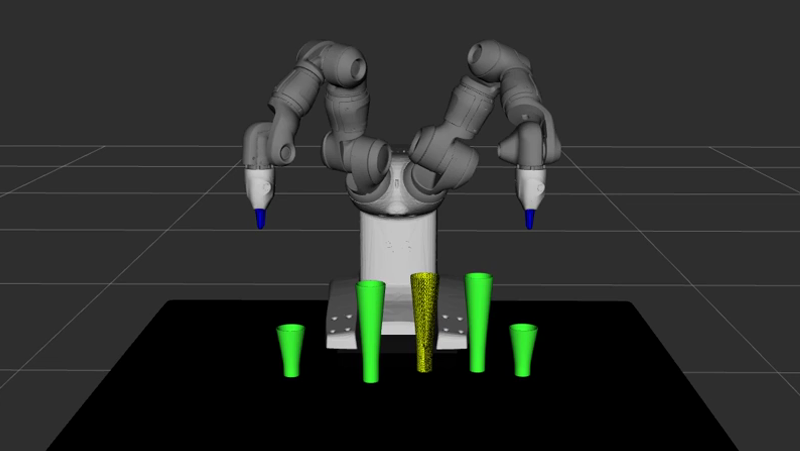}
    \vspace{-17pt}
    \caption{}
  \end{subfigure}\hspace{-2pt}
  \begin{subfigure}[b]{.497\linewidth}
    \includegraphics[width=\textwidth]{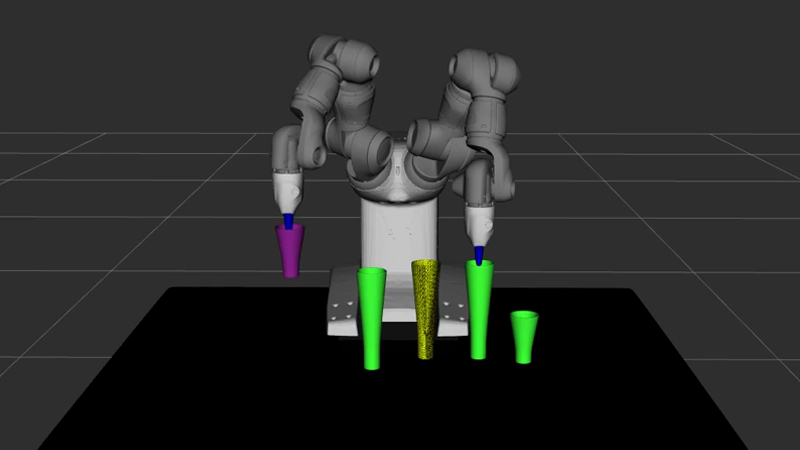}
    \vspace{-17pt}
    \caption{}
  \end{subfigure}\\\vspace{0.5pt}
  \begin{subfigure}[b]{.497\linewidth}
    \includegraphics[width=\textwidth]{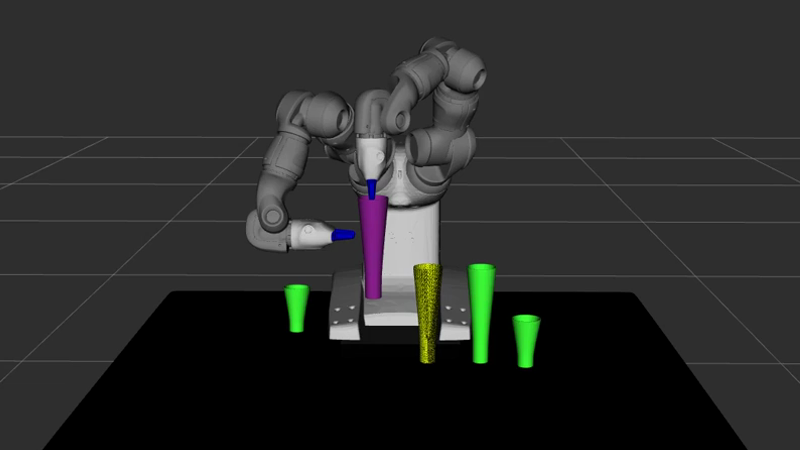}
    \vspace{-17pt}
    \caption{}
  \end{subfigure}\hspace{-2pt}
  \begin{subfigure}[b]{.497\linewidth}
    \includegraphics[width=\textwidth]{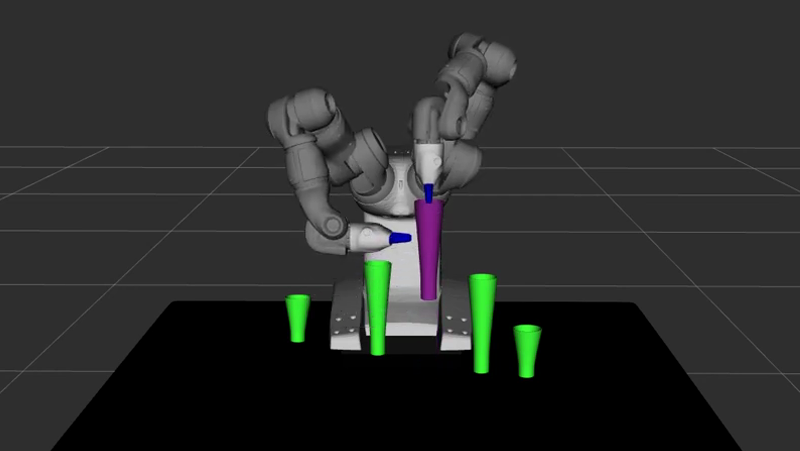}
    \vspace{-17pt}
    \caption{}
  \end{subfigure}
\caption{Snapshots of performing a dualarm tabletop object 
task. (a) the initial state; (b) and (c) removing obstructing
objects from the table to unblock the left grasp pose of the 
target object, after 9 replanning and feasibility testing; (d) 
the task is achieved.}
\label{fig:exp:tabletop}
\end{center}
\end{figure}

\paragraph{Dualarm Tabletop Object Manipulation}\label{sec:tabletop}
The second task involves picking up an object where its 
grasp poses might be obstructed by neighboring objects (see 
Figure~\ref{fig:exp:tabletop}).
The domain allows for picking up an object with both 
single and dualarm actions, however, a strong solution 
to pick up a large object must involve two arms rather 
than a single arm since the object might fall down and slip 
from the table, resulting in a deadend state. Moreover, the 
obstruction constraints are not known in advance by the robot.
In this experiment, the TAMP framework computed a feasible 
policy to achieve the task after testing the feasibility 
of 24 actions, identifying 9 unfeasible actions (constraints), 
and making 9 replannings in under 25 seconds.

